# High-dimensional dynamics of generalization error in neural networks

**Madhu S. Advani**[*]     madvani@fas.harvard.edu
*Center for Brain Science*
*Harvard University*
*Cambridge, MA 02138*

**Andrew M. Saxe**[*]     asaxe@fas.harvard.edu
*Center for Brain Science*
*Harvard University*
*Cambridge, MA 02138*

**Editor:** TBD

## Abstract

We perform an average case analysis of the generalization dynamics of large neural networks trained using gradient descent. We study the practically-relevant "high-dimensional" regime where the number of free parameters in the network is on the order of or even larger than the number of examples in the dataset. Using random matrix theory and exact solutions in linear models, we derive the generalization error and training error dynamics of learning and analyze how they depend on the dimensionality of data and signal to noise ratio of the learning problem. We find that the dynamics of gradient descent learning naturally protect against overtraining and overfitting in large networks. Overtraining is worst at intermediate network sizes, when the effective number of free parameters equals the number of samples, and thus can be reduced by making a network smaller or *larger*. Additionally, in the high-dimensional regime, low generalization error requires starting with small initial weights. We then turn to non-linear neural networks, and show that making networks very large does not harm their generalization performance. On the contrary, it can in fact reduce overtraining, even without early stopping or regularization of any sort. We identify two novel phenomena underlying this behavior in overcomplete models: first, there is a frozen subspace of the weights in which no learning occurs under gradient descent; and second, the statistical properties of the high-dimensional regime yield better-conditioned input correlations which protect against overtraining. We demonstrate that naive application of worst-case theories such as Rademacher complexity are inaccurate in predicting the generalization performance of deep neural networks, and derive an alternative bound which incorporates the frozen subspace and conditioning effects and qualitatively matches the behavior observed in simulation.

## 1. Introduction

Deep learning approaches have attained high performance in a variety of tasks (LeCun, Bengio, and Hinton, 2015; Schmidhuber, 2015), yet their learning behavior remains opaque.

---

[*]. Both authors contributed equally to this work.

Strikingly, even very large networks can attain good generalization performance from a relatively limited number of examples (Canziani, Paszke, and Culurciello, 2017; He et al., 2016; Simonyan and Zisserman, 2015). For instance, the VGG deep object recognition network contains 155 million parameters, yet generalizes well when trained with just the 1.2 million examples of ImageNet (Simonyan and Zisserman, 2015). This observation leads to important questions. What theoretical principles explain this good performance in this limited-data regime? Given a fixed dataset size, how complex should a model be for optimal generalization error?

In this paper we study the average generalization error dynamics of various simple training scenarios in the "high-dimensional" regime where the number of samples $P$ and parameters $N$ are both large ($P, N \to \infty$), but their ratio $\alpha = P/N$ is finite (e.g Advani, Lahiri, and Ganguli, 2013). We start with simple models, where analytical results are possible, and progressively add complexity to verify that our findings approximately hold in more realistic settings. We mainly consider a student-teacher scenario, where a "teacher" neural network generates possibly noisy samples for a "student" network to learn from (Seung, Sompolinsky, and Tishby, 1992; Watkin, Rau, and Biehl, 1993; Saad and Solla, 1995). This simple setting retains many of the trade-offs seen in more complicated problems: under what conditions will the student overfit to the specific samples generated by the teacher network? How should the complexity (e.g. number of hidden units) of the student network relate to the complexity of the teacher network?

First, in Section 2, we investigate linear neural networks. For shallow networks, it is possible to write exact solutions for the dynamics of batch gradient descent learning as a function of the amount of data and signal to noise ratio by integrating random matrix theory results and the dynamics of learning. This is related to a large body of prior work on shallow networks (Chauvin, 1990; P. Baldi and Chauvin, 1991; LeCun, Kanter, and Solla, 1991; Seung, Sompolinsky, and Tishby, 1992; Dodier, 1995; P.F. Baldi and Hornik, 1995; Kinouchi and Caticha, 1995; Dunmur and Wallace, 1999; Krogh and Hertz, 1992; Hoyle and Rattray, 2007; Bai and Silverstein, 2010; Benaych-Georges and Rao, 2011; Benaych-Georges and Rao, 2012). Our results reveal that the combination of early stopping and initializing with small-norm weights successfully combats overtraining, and that overtraining is worst when the number of samples equals the number of model parameters.

Interestingly the optimal stopping time (for minimizing generalization error) in these systems growths with the signal to noise ratio (SNR), and we provide arguments that this growth is approximately logarithmic. When the quality of the data is higher (high SNR), we expect the algorithm to weigh the data more heavily and thus run gradient descent for longer. We also analyze the training error dynamics and demonstrate that the time required to approximately minimize the training error will depend weakly on the signal to noise ratio of the problem, typically growing logarithmically with one over the SNR. Thus, noisier data will take slightly longer to achieve a low training error.

For deep linear neural networks, in Section 3 we derive a reduction of the full coupled, nonlinear gradient descent dynamics to a much smaller set of parameters coupled only through a global scalar. Our reduction is applicable to any depth network, and yields insight into the dynamical impact of depth on training time. The derivation here differs from previous analytic works on the dynamics of deep linear networks (Saxe, McClelland, and Ganguli, 2014) in that we do not assume simultaneous diagonalizability of the input and input-output correlations.

Next, in Section 4 we turn to nonlinear networks. We consider a nonlinear student network which learns from a dataset generated by a nonlinear teacher network. We show through simulation that the qualitative intuitions gained from the linear analysis transfer well to the nonlinear setting. Remarkably, we find that catastrophic overtraining is a symptom of a model whose complexity is exactly matched to the size of the training set, and can be combated either by making the model smaller or *larger*. Moreover, we find no evidence of overfitting: the optimal early-stopping generalization error decreases as the student grows larger, even when the student network is much larger than the teacher and contains many more hidden neurons than the number of training samples. Our findings agree both with early numerical studies of neural networks (Caruana, Lawrence, and Giles, 2001) and the good generalization performance of massive neural networks in recent years (e.g. Simonyan and Zisserman, 2015). In this setting, we find that bigger networks are better, and the only cost is the computational expense of the larger model.

We also analyze a two layer network to gain theoretical intuition for these numerical results. A student network with a wide hidden layer can be highly expressive (Barron, 1993), even if we choose the first layer to be random and fixed during training. However, in this setting, when the number of hidden units $N_h$ is larger than the number of samples $P$, our linear solution implies that there will be no learning in the $N_h - P$ zero-eigenvalue directions of the hidden layer covariance matrix. It follows that further increasing the network size will not actually increase the complexity of the functions the network can learn. Furthermore, there seems to be an implicit reduction in model complexity and overfitting at late stopping times when $N_h$ is increased above $P$ because the non-zero eigenvalues are actually pushed away from the origin, thus reducing the maximum norm learned by the student weights. Finally, through simulations with the MNIST dataset, we show that these qualitative phenomena are recapitulated on real-world datasets.

The excellent performance of complex models may seem to contradict straightforward applications of VC dimension and Rademacher bounds on generalization performance (C. Zhang et al., 2017). We take up these issues in Sections 5 and 6. Our results show that the effective complexity of the model is regularized by the combined strategy of early stopping and initialization with small norm weights. This two-part strategy limits the norm of the weights in the network, thereby limiting the Rademacher complexity. This strategy is analogous to that used in support vector machines to allow generalization from kernels with infinite VC dimension: generalization is preserved provided the training process finds a large-margin (low-norm weight vector) solution on the training data. Hence, while a large deep network can indeed fit random labels, gradient-trained DNNs initialized with small-norm weights learn simpler functions first and hence generalize well if there is a consistent rule to be learned.

Remarkably, even without early stopping, generalization can still be better in larger networks. To understand this, we derive an alternative bound on the Rademacher complexity of a two layer non-linear neural network with fixed first layer weights that incorporates the dynamics of gradient descent. We show that complexity is limited because of a frozen subspace in which no learning occurs, and overtraining is prevented by a larger gap in the eigenspectrum of the data in the hidden layer in overcomplete models. Our bound provides new intuitions for generalization behavior in large networks, and qualitatively matches the findings of our simulations and prior work: even without early stopping, overtraining is reduced by increasing the number of hidden units $N_h$ when $N_h > P$.

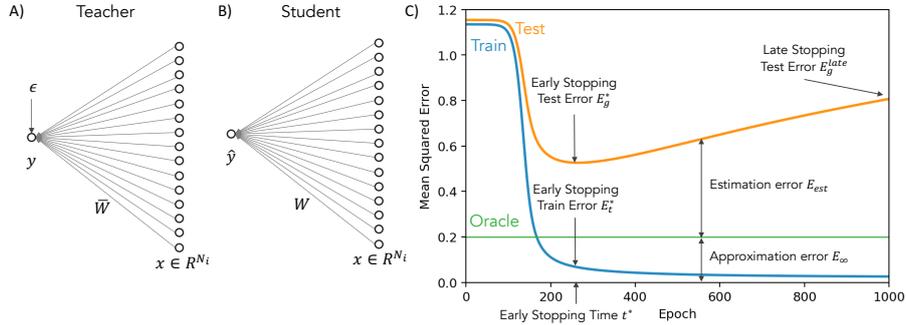

Figure 1: Learning from a noisy linear teacher. A) A dataset $\mathcal{D} = \{x^\mu, y^\mu\}, \mu = 1, \cdots, P$ of $P$ examples is created by providing random Gaussian inputs $x$ to a teacher network with randomly drawn parameters $\bar{w}$, and corrupting these with Gaussian noise of variance $\sigma_\epsilon^2$. B) A student network is then trained on this dataset $\mathcal{D}$. C) Example dynamics of the student network during full batch gradient descent training. Training error (blue) decreases monotonically. Test error (yellow), here computable exactly, decreases to a minimum $E_g^*$ at the optimal early stopping time $t^*$ before increasing at longer times ($E_g^{\text{late}}$), a phenomenon known as overtraining. Because of noise in the teacher's output, the best possible student network attains finite generalization error ("oracle", green) even with infinite training data. This error is the approximation error $E_\infty$. The difference between test error and this best-possible error is the estimation error $E_{\text{est}}$.

Our results begin to shed light on the diverse learning behaviors observed in deep networks, and support an emerging view that suggests erring on the side of large models before pruning them later.

## 2. Generalization dynamics in shallow linear neural networks

To begin, we study the simple case of generalization dynamics in a shallow linear neural network receiving $N$-dimensional inputs. We consider a standard student-teacher formulation, a setting that has been widely studied using statistical mechanics approaches (see Seung, Sompolinsky, and Tishby, 1992; Watkin, Rau, and Biehl, 1993; Engel and Van Den Broeck, 2001 for reviews, and P. Baldi, Chauvin, and Hornik, 1990; Chauvin, 1990; P. Baldi and Chauvin, 1991; LeCun, Kanter, and Solla, 1991; P.F. Baldi and Hornik, 1995; Kinouchi and Caticha, 1995; Dunmur and Wallace, 1999 for settings more closely related to the linear analysis adopted in this section). Here a student network with weight vector $w(t) \in \mathbb{R}^{1 \times N}$ trains on examples generated by a teacher network with weights $\bar{w} \in \mathbb{R}^{1 \times N}$ (see Fig. 1). The teacher implements a noisy linear mapping between $P$ inputs $X \in \mathbb{R}^{N \times P}$ and associated scalar outputs $y \in \mathbb{R}^{1 \times P}$,

$$y = \bar{w}X + \epsilon. \tag{1}$$

In the equation above, $\epsilon \in \mathbb{R}^{1 \times P}$ denotes noise in the teacher's output. We will model both the noise $\epsilon$ and the teacher weights $\bar{w}$ as drawn *i.i.d.* from a random Gaussian distribution with zero mean and variance $\sigma_\epsilon^2$ and $\sigma_w^2$ respectively. In this setting, the function to be learned ($\bar{w}$) is a randomly drawn mapping from input to output; this is a simple stand in for learning some task of interest. The signal-to-noise ratio $\text{SNR} \equiv \sigma_w^2/\sigma_\epsilon^2$ parametrizes the strength of the rule underlying the dataset relative to the noise in the teacher's output. In our solutions, we will compute the generalization dynamics averaged over all possible realizations of $\bar{w}$ to access the general features of the learning dynamics independent of the specific problems encountered. Finally, we assume that the inputs $X_{\mu j}$ are drawn *i.i.d.* from a Gaussian with mean zero and variance $\frac{1}{N}$ so that each example will have an expected norm of one: $\left\langle \|x^\mu\|_2^2 \right\rangle = 1$.

The student network is trained using the dataset $\{y, X\}$ to accurately predict outputs for novel inputs $x \in \mathbb{R}^N$. The student is a shallow linear network, such that the student's prediction $\hat{y} \in \mathbb{R}$ is simply $\hat{y}^\mu = w(t)x^\mu$. To learn its parameters, the student network will attempt to minimize the mean squared error on the $P$ training samples using gradient descent. The training error is

$$E_t(w(t)) = \frac{1}{P}\sum_{\mu=1}^{P}\|y^\mu - \hat{y}^\mu\|_2^2, \tag{2}$$

and full-batch continuous-time gradient descent on this error function (a good approximation to backpropagation with small step size) yields the dynamical equations

$$\tau \dot{w}(t) = yX^T - wXX^T. \tag{3}$$

where $\tau$ is a time constant inversely proportional to the learning rate. Here the time variable $t$ tracks the number of epochs, such that as $t$ goes from 0 to 1, for example, the student network has seen all $P$ examples once. Our primary goal is to understand the evolving generalization performance of the network over the course of training. That is, we wish to understand the generalization error on a new, unseen example,

$$E_g(w(t)) = \left\langle (y - \hat{y})^2 \right\rangle_{x,\epsilon} \tag{4}$$

as training proceeds. Here the average $\langle \cdot \rangle_{x,\epsilon}$ is over potential inputs $x$ and noise $\epsilon$.

We emphasize that we are analyzing the batch setting, in which a fixed set of $P$ examples is used again and again throughout training. This contrasts with the online setting, where each example is used only once. In the online setting, overtraining is impossible as each new example yields an unbiased sample from the true generalization error gradient (see e.g., Fig. 11 of Erhan et al., 2010 for an empirical demonstration in a deep neural network). Overtraining, thus, is a phenomenon of batch learning. In practice, neural networks are typically trained with stochastic gradient descent (SGD) such that weight updates are applied after each example, which may seem closer to the online setting. However, typical training procedures based on SGD make multiple passes through the same fixed dataset, and this is the essential feature of batch learning: repeatedly training on the same examples permits overtraining to the idiosyncrasies of the particular batch. Hence our analysis is a reasonable description of SGD dynamics if examples are revisited many times and the learning rate is small.

**2.1 Exact solutions in the high-dimensional limit**

How does generalization performance evolve over the course of training? The long-term behavior is clear: if training is run for a long time, the weight vector $w$ will converge to the subspace minimizing the training error (2), solving the least-squares regression problem,

$$w(t \to \infty) = yX^T(XX^T)^+, \qquad (5)$$

where $+$ denotes a pseudoinverse. The error achieved by the pseudoinverse, along with more general results for nonlinear students and teachers, have long been known (LeCun, Kanter, and Solla, 1991; Seung, Sompolinsky, and Tishby, 1992), we explicitly compute the time course with which parameters are learned.

First, we decompose the input correlation matrix using the eigendecomposition,

$$\Sigma^{xx} = XX^T = V\Lambda V^T. \qquad (6)$$

Next we write the input-output correlation matrix as

$$\Sigma^{yx} \;=\; yX^T = \tilde{s}V^T, \qquad (7)$$

where the row vector $\tilde{s}$ will be called an alignment vector since it is related to the alignment between the input-output correlations and the input correlations.

Returning to the dynamics in (3), we make a change of variables to instead track the vector $z \in \mathbb{R}^{1 \times N}$ where $w = zV^T$. Applying (7) yields:

$$\tau \dot{z}(t) = \tilde{s} - z\Lambda. \qquad (8)$$

To make sense of the preceding equation we compute $\tilde{s}$ using the fact that

$$yX^T = \bar{w}XX^T + \epsilon X^T = \bar{z}\Lambda V^T + \tilde{\epsilon}\Lambda^{1/2}V^T, \qquad (9)$$

where we define $\bar{z} = \bar{w}V$, and under the assumption of white Gaussian noise, $\tilde{\epsilon} \in \mathbb{R}^N$ has *i.i.d.* elements drawn from a Gaussian with variance $\sigma_\epsilon^2$. The form of $\Sigma^{yx}$ implies $\tilde{s} = \bar{z}\Lambda + \tilde{\epsilon}\Lambda^{1/2}$, so we can rewrite (8) as

$$\tau \dot{z}_i = (\bar{z}_i - z_i)\lambda_i + \tilde{\epsilon}_i\sqrt{\lambda_i}, \qquad i = 1, \cdots, N. \qquad (10)$$

The learning speed of each mode is independent of the others and depends only on the eigenvalue of the mode in question. As we will see, in the case of deep learning, there will be coupling between these modes. However in shallow neural networks, such coupling does not occur and we can solve directly for the dynamics of these modes. The error in each component of $z$ is simply

$$\bar{z}_i - z_i = (\bar{z}_i - z_i(0))e^{-\frac{\lambda_i t}{\tau}} - \frac{\tilde{\epsilon}_i}{\sqrt{\lambda_i}}(1 - e^{-\frac{\lambda_i t}{\tau}}). \qquad (11)$$

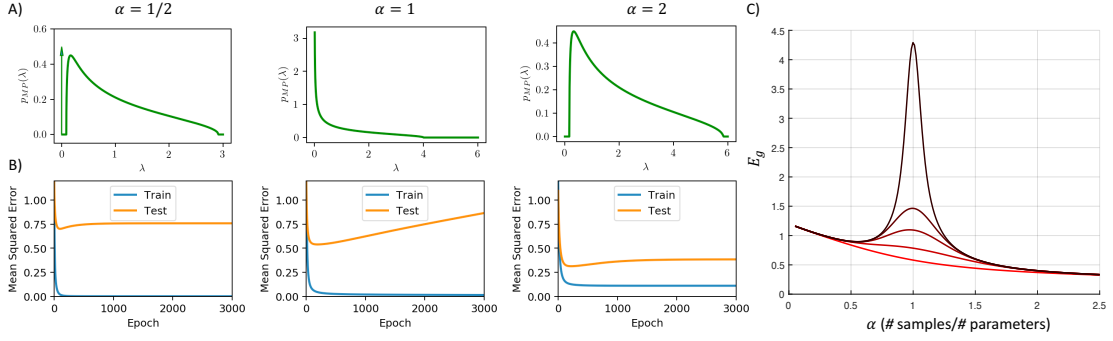

Figure 2: The Marchenko-Pasteur distribution and high-dimensional learning dynamics. A) Different ratios of number training samples ($P$) to network parameters ($N$) ($\alpha = \frac{P}{N}$) yield different eigenvalue densities in the input correlation matrix. For large $N$, this density is described by the MP distribution (14), which consists of a 'bulk' lying between $[\lambda_-, \lambda_+]$, and, when $\alpha < 1$, an additional delta function spike at zero. When there are fewer samples than parameters ($\alpha < 1$, left column), some fraction of eigenvalues are exactly zero (delta-function arrow at origin), and the rest are appreciably greater than zero. When the number of samples is on the order of the parameters ($\alpha = 1$, center column), the distribution diverges near the origin and there are many nonzero but arbitrarily small eigenvalues. When there are more samples than parameters ($\alpha > 1$, right column), the smallest eigenvalues are appreciably greater than zero. B) Dynamics of learning. From (13), the generalization error is harmed most by small eigenvalues; and these are the slowest to be learned. Hence for $\alpha = 1/2$ and $\alpha = 2$, the gap in the spectrum near zero protects the dynamics from overtraining substantially (eigenvalues which are exactly zero for $\alpha = 1/2$ are never learned, and hence contribute a finite error but no overtraining). For $\alpha = 1$, there are arbitrarily small eigenvalues, and overtraining is substantial. C) Plot of generalization error versus $\alpha$ for several training times, revealing a clear spike near $\alpha = 1$. Other parameters: $N = 100, \text{INR} = 0, \text{SNR} = 5$. As the colors vary from red to black the training time increases $\frac{t}{\tau} = [5, 20, 50, 100, 1000]$.

It follows from the definition of $z$ that the generalization error as a function of training time is

$$E_g(t) = \frac{1}{N}\sum_i \left\langle (\bar{z}_i - z_i)^2 \right\rangle + \sigma_\epsilon^2 \tag{12}$$

$$= \frac{1}{N}\sum_i \left[(\sigma_w^2 + (\sigma_w^0)^2)e^{-\frac{2\lambda_i t}{\tau}} + \frac{\sigma_\epsilon^2}{\lambda_i}(1 - e^{-\frac{\lambda_i t}{\tau}})^2\right] + \sigma_\epsilon^2, \tag{13}$$

where the second equality follows from the assumption that the teacher weights $\bar{w}$ and initial student weights $w(0)$ are drawn *i.i.d.* from Gaussian distributions with standard deviation $\sigma_w$ and $\sigma_w^0$ respectively. The generalization error expression contains two time-dependent terms. The first term exponentially decays to zero, and encodes the distance between the weight initialization and the final weights to be learned. The second term begins at zero and exponentially approaches its asymptote of $\frac{\sigma_\epsilon^2}{\lambda_i}$. This term corresponds to overfitting the noise present in the particular batch of samples. We note two important points: first, eigenvalues which are exactly zero ($\lambda_i = 0$) correspond to directions with no learning dynamics so that the parameters $z_i$ will remain at $z_i(0)$ indefinitely. These directions form a *frozen subspace* in which no learning occurs. Hence, if there are zero eigenvalues, weight initializations can have a lasting impact on generalization performance even after arbitrarily long training. Second, smaller eigenvalues lead to the most serious over-fitting due to the $\frac{\sigma_\epsilon^2}{\lambda_i}$ factor in the second term of the generalization error expression. Hence a large *eigengap* between zero and the smallest nonzero eigenvalue can naturally protect against overfitting. Moreover, smaller eigenvalues are also the slowest to learn, suggesting that early stopping can be an effective strategy, as we demonstrate in more detail subsequently. Finally, the expression provides insight into the time required for convergence. Non-zero but small eigenvalues of the sample covariance lead to very slow dynamics, so that it will take on the order of $t = \frac{\tau}{\lambda_{\min}}$ for gradient descent to minimize the training error.

The result in (13) reveals the critical role played by the eigenvalue spectrum of the sample input covariance matrix. If there are many small eigenvalues generalization performance will be poor, while if there are only large eigenvalues generalization performance will be strong. What is the distribution of this spectrum as a function of the signal-to-noise ratio and size of the dataset? To understand this, we pass to the high-dimensional limit where the input dimension $N$ and number of examples $P$ jointly go to infinity, while their ratio $\alpha = P/N$ remains finite. Then the eigenvalue distribution of $XX^T$ approaches the Marchenko-Pasteur distribution (Marchenko and Pasteur, 1967; LeCun, Kanter, and Solla, 1991),

$$\rho^{\text{MP}}(\lambda) = \frac{1}{2\pi}\frac{\sqrt{(\lambda_+ - \lambda)(\lambda - \lambda_-)}}{\lambda} + 1_{\alpha<1}(1-\alpha)\delta(\lambda), \tag{14}$$

for $\lambda = 0$ or $\lambda \in [\lambda_-, \lambda_+]$, and the distribution is zero elsewhere. Here the edges of the distribution take the values $\lambda_\pm = (\sqrt{\alpha} \pm 1)^2$ and hence depend on the number of examples relative to the input dimension. Figure 2A depicts this distribution for three different values of the load $\alpha$. In the undersampled regime when $\alpha < 1$, there are fewer examples than input dimensions and many eigenvalues are exactly zero, yielding the delta function at the origin. This corresponds to a regime where data is scarce relative to the size of the model. In the critically sampled regime $\alpha = 1$, there are exactly as many parameters as examples. Here the distribution diverges towards the origin, with increasing probability of extremely small eigenvalues. This situation yields catastrophic overtraining. Finally in the oversampled regime when $\alpha > 1$, there are more examples than input dimensions, yielding the traditional asymptotic regime where data is abundant. The eigenvalue distribution in this case is shifted away from the origin.

Combining the dynamical solution (13) with the spectrum distribution (14) yields the predicted typical generalization error,

$$\frac{E_g(t)}{\sigma_w^2} = \int \rho^{\text{MP}}(\lambda) \left[ (1 + \text{INR})e^{-\frac{2\lambda t}{\tau}} + \frac{1}{\lambda \cdot \text{SNR}}(1 - e^{-\frac{\lambda t}{\tau}})^2 \right] d\lambda + \frac{1}{\text{SNR}}, \tag{15}$$

where we have normalized by $\sigma_w^2$ to set the scale and defined the initialization noise ratio $\text{INR} \equiv (\sigma_w^0)^2 / \sigma_w^2$. Figure 2B shows the resulting generalization dynamics for the under-, critically- and over-sampled regimes. While all three exhibit overfitting, this is substantially worse at the intermediate point $\alpha = 1$. Fig. 2C systematically traces out generalization performance as a function of $\alpha$ at several training times, showing a clear divergence at $\alpha = 1$. Hence overtraining can lead to complete failure of generalization performance when the measurement density $\alpha$ is close to 1. The generalization error prediction (15) is also validated in Fig 3 where we compare theory with simulations.

Remarkably, even this simple setting of a linear student and teacher yields generalization dynamics with complex overtraining phenomena that depend on the parameters of the dataset. In the following sections we explore aspects of these solutions in greater detail, and show that early stopping provides an effective remedy to combat overtraining.

### 2.2 Effectiveness of early stopping vs L2 regularization

The generalization curves in Fig. 2B improve for a period of time before beginning to worsen and converge to the performance of $w(\infty)$. Thus, there will be an optimal stopping time at which simply ending training early would yield improved generalization compared to training for longer times. The intuitive explanation for this is that by limiting time we are effectively regularizing the parameters being learned. This early stopping strategy is widely used in practice, and has been studied theoretically (Chauvin, 1990; P.F. Baldi and Hornik, 1995; Dodier, 1995; Amari, Murata, Müller, et al., 1995; Amari, Murata, and Müller, 1996; Biehl, Caticha, and Riegler, 2009; Yao, Rosasco, and Caponnetto, 2007). Here we compare the performance of early stopping of gradient descent to the performance of L2 regularization, which corresponds to solving the optimization problem:

$$\hat{w}_{L2} = \arg\min_w \frac{1}{2} \left[ \|y - wX\|_2^2 + \frac{\gamma}{2} \|w\|_2^2 \right]. \tag{16}$$

The optimal performance of this algorithm occurs when the regularization strength is tuned to be inversely proportional to the signal-to-noise ratio: $\gamma = \frac{1}{\text{SNR}} = \frac{\sigma_\epsilon^2}{\sigma_w^2}$. In fact, under the assumptions we make of Gaussian noise and parameter distributions, no algorithm can outperform optimal L2 regularization in terms of generalization error performance (as shown in e.g. Advani and Ganguli, 2016a; Advani and Ganguli, 2016c). Here we ask how close the performance of early stopping comes to this best-possible performance. We compare the two algorithms in Fig. 4A. There is a very close match between their generalization performances: here the relative error between the two is under 3 percent and is the largest around $\alpha = 1$. Hence early stopping can be a highly effective remedy against overtraining. Further, we can exploit the similar performance of these two algorithms to make analytic predictions for how the optimal stopping time depends on other parameters, namely the measurement density and SNR.

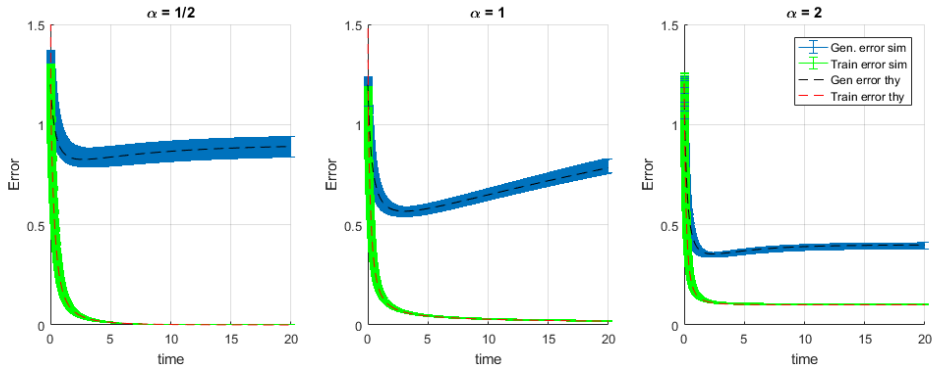

Figure 3: Generalization and training error dynamics compared with simulations. Here we demonstrate that our theoretical predictions for generalization and training error dynamics (dashed lines) at different measurement densities ($\alpha = 1/2, \alpha = 1$, and $\alpha = 2$) match with simulations. The simulations for generalization (blue) and training (green) errors have a width of $\pm 2$ standard deviations generated from 20 trials using $N = 300$, $P = \alpha N$ and Gaussian input, noise, and parameters with $\sigma_w = 1, \sigma_w^0 = 0$, and SNR $= 5$. The simulations show excellent agreement with our theoretical predictions for generalization and training error provided in (15) and (27).

### 2.3 Optimal stopping time vs SNR

As discussed earlier, training to convergence requires training for time proportional to $1/\lambda_{min}$, the smallest eigenvalue. The optimal stopping time, however, can be substantially shorter. In this section we estimate its dependence on the parameters of the problem. If we initialize with $w(0) = 0$, the expression for $E_g(t)$ reduces to

$$\frac{E_g(t)}{\sigma_w^2} = \int \rho^{\mathrm{MP}}(\lambda) \left[ e^{-\frac{2\lambda t}{\tau}} + \frac{1}{\lambda \cdot \mathrm{SNR}} (1 - e^{-\frac{\lambda t}{\tau}})^2 \right] d\lambda + \frac{1}{\mathrm{SNR}}. \tag{17}$$

To solve for the optimal stopping time numerically, we can differentiate the above equation with respect to $t$ and set the result equal to zero. However, to gain insight into how the optimal stopping time depends on measurement density and SNR, it is helpful to compare it to L2-regularized regression where the generalization error (see Advani and Ganguli, 2016b)

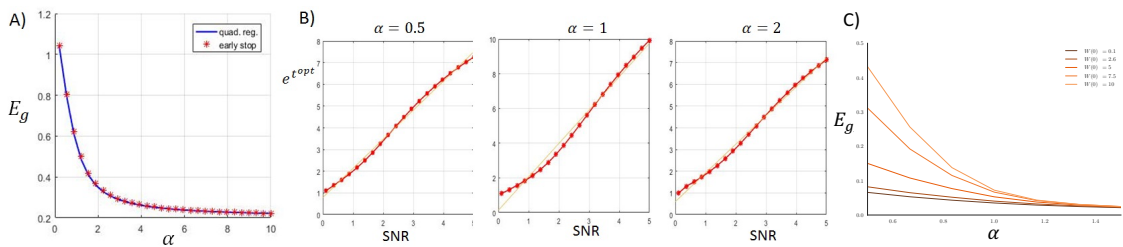

Figure 4: Comparison of early stopping to optimal quadratic (L2) regularization in shallow networks. In (A) we see that the optimal early stopping prediction (red stars) achieves a generalization error near that of optimal quadratic regularization (blue line). In this example there is no more than 3 percent relative error between the two, which peaks near $\alpha = 1$ when the spectrum of non-zero eigenvalues is broader. Here we have fixed SNR = 5 and $\sigma_w^2 = 1$ which implies that even at large measurement density $\alpha$ the generalization error asymptotes to a non-zero value due to noise $0.2 = \frac{1}{\text{SNR}}$. (B) The approximation for optimal stopping time in (19) correctly predicts the scaling of optimal stopping time with SNR. $\alpha$ moderates the slope of the effect of SNR on the optimal stopping time by shifting where the bulk of the eigenvalues rest. Note that the case of $\alpha = 1$ has a large spread of eigenvalues so that the assumptions we make to derive the scaling are weakest in this limit leading to a slightly worse fit. (C) Generalization error vs $\alpha$ for different initial weight norms. In the limited data regime, small initial weights are crucial for good generalization.

has the form:

$$\frac{E_g^{L2}(\gamma)}{\sigma_w^2} = \int \rho^{\text{MP}}(\lambda) \left[ \frac{\gamma^2}{(\lambda+\gamma)^2} + \frac{\lambda^2}{\lambda \cdot \text{SNR}(\lambda+\gamma)^2} \right] d\lambda + \frac{1}{\text{SNR}}. \tag{18}$$

Selecting the optimal regularization strength $\gamma = \frac{1}{\text{SNR}}$ and equating the terms inside the square brackets in the two preceding generalization errors allows us to compute the time at which any given mode will be regularized to the optimal level attained by quadratic regularization. For a mode with eigenvalue $\lambda$ this occurs at:

$$t^{opt} = \frac{\tau}{\lambda} \log(\text{SNR} \cdot \lambda + 1). \tag{19}$$

Thus, if the eigenspectrum were concentrated on one value equal to $\lambda$, optimal L2 regularization performance would be achieved at the optimal stopping time above. Because the non-zero eigenvalues are spread, particularly when $\alpha$ is near 1, they will not all reach the optimal level of regularization at the same time; this is the source of the slight suboptimality in the early stopping procedure. However, this approximation does surprisingly give a reasonably accurate scaling of optimal stopping time with SNR as shown in Fig 4B: a linear fit of the exponential of early stopping time is a good match here, and is worst around $\alpha = 1$. The intuition behind the logarithmic growth of optimal stopping time with SNR is that high quality data requires less regularization. In other words, the reduced danger of overtraining justifies running gradient descent for longer.

### 2.4 Weight norm growth with SNR

Given that longer training times provide the opportunity for more weight updates, one might also expect that the norm of the weights at the optimal early stopping time would vary systematically with SNR, and we find this is indeed the case. Assuming zero initial student weights, the norm of the optimal quadratic regularized weights (derived in Appendix B) increases with SNR as:

$$\langle \hat{w}_{L2}^2 \rangle = \sigma_w^2 \int \rho^{MP}(\lambda) \left( \frac{\lambda}{\lambda + \frac{1}{\text{SNR}}} \right) d\lambda. \tag{20}$$

The intuition behind this equation is that data with a lower SNR requires greater regularization which leads to a lower norm in the learned weights. The norm of the learned weights interpolates between $\langle \hat{w}^2 \rangle \to \sigma_w^2$ as $\text{SNR} \to \infty$ and $\langle \hat{w}^2 \rangle \to 0$ as $\text{SNR} \to 0$. Thus the signal-to-noise ratio of the data plays a pivotal role in determining the norm of the learned weights at the optimal stopping time.[1]

---

1. This dependence of weight norm on SNR may help reconcile puzzles in recent empirical papers. For instance, Section 5 of C. Zhang et al., 2017 compares linear maps learned off of two different nonlinear transformations of the input, and finds that the linear map with larger norm sometimes generalizes better–in aparent contradiction to generalization bounds that rely on norm-based capacity control. However, based on our results, this finding is expected if one nonlinear transformation reveals more signal than the other: the transformation yielding higher SNR will require a longer training time, yield larger norm weights, and attain better generalization performance. Here the larger norm weights are justified by the stronger rule linking input and output. Comparisons between linear maps of different norms learned from the same nonlinear transformation, however, will reveal an optimal weight norm (which can be approximately found through early stopping).

### 2.5 Impact of initial weight size on generalization performance

In the preceding two sections, we have shown the effectiveness of early stopping provided that the student is initialized with zero weights (INR $\approx 0$). The early stopping strategy leaves weights closer to their initial values, and hence will only serve as effective regularization when the initial weights are small. To elucidate the impact of weight initialization on generalization error, we again leverage the similar performance of early stopping and L2 regularization. If the student weights are not initially zero but instead have variance $(\sigma_w^0)^2$, the optimal L2 regularization strength is $\gamma^{\text{opt}} = \frac{\sigma_\epsilon^2}{(\sigma_w^0)^2 + \sigma_w^2}$. This yields the generalization error,

$$E_g = \frac{(\sigma_w^0)^2 + \sigma_w^2}{2} \left(1 - \alpha - \gamma^{\text{opt}} + \sqrt{(\gamma^{\text{opt}} + \alpha - 1)^2 + 4\gamma^{\text{opt}}}\right) + \sigma_\epsilon^2. \tag{21}$$

In the limit of high SNR, there are three qualitatively different behaviors for the dependence of generalization performance on the initial weight values in the under-sampled, equally sampled, and oversampled regimes. There is a linear dependence on initial weight size when $\alpha < 1$, a square root dependence for $\alpha = 1$, and no dependence when $\alpha > 1$,

$$E_g \approx \begin{cases} ((\sigma_w^0)^2 + \sigma_w^2)(1 - \alpha) + \sigma_\epsilon^2, & \alpha < 1 \\ \sqrt{(\sigma_w^0)^2 + \sigma_w^2}\sigma_\epsilon + \sigma_\epsilon^2, & \alpha = 1 \\ \sigma_\epsilon^2 \frac{\alpha}{\alpha - 1}, & \alpha > 1. \end{cases}$$

Hence as more data becomes available, the impact of the weight initialization decreases. This reflects the decreasing utility of regularization when data is plentiful, or equivalently, the reduced influence of a prior after many observations. Our dynamical solutions reveal the source of this effect. When there are few examples relative to the input dimension, there are many directions in the input space along which weights do not change because no examples lie in those directions. Hence in this frozen subspace, the weights will remain at their initial values indefinitely. In particular, the frozen subspace is spanned by the eigenvectors of the input covariance matrix which have zero eigenvalues, and there are a fraction $1 - \alpha$ of these for $\alpha < 1$. Therefore, in the high-dimensional regime where data is limited relative to the size of the model, it is critical to initialize with small weights to maximize generalization performance (see Fig. 4C). Even when the number of examples is matched to the size of the model ($\alpha \approx 1$), large-norm weight initializations remain detrimental for the optimal early stopping strategy because training must terminate before the influence of the initial weights has fully decayed. Hence, based on this simple linear analysis, excellent generalization performance in large models can be obtained through the two-part strategy of early stopping and initialization with small-norm weights. Remarkably, this strategy nearly matches the performance of the best posssible Bayes-optimal learning strategy (optimal L2 regularization) for the setting we consider, and remains effective in the high-dimensional regime where the number of samples can be smaller than the size of the model.

### 2.6 Training error dynamics

Our theory also allows us to predict the dynamics of the training error during learning. As we will show, these dynamics depend strongly on the small eigenvalues of the data covariance and weakly on the SNR. To derive the training error dynamics, we begin with the form:

$$E_t(w(t)) = \frac{1}{P}\|y - w(t)X\|_2^2. \tag{22}$$

Substituting the singular value decomposition of the data $X = V\Lambda^{1/2}U^T$ (with $U \in \mathbb{R}^{P \times N}, \Lambda^{1/2} \in \mathbb{R}^{N \times N}$, and $V \in \mathbb{R}^{N \times N}$) and $w(t) = z(t)V^T$ yields:

$$\frac{1}{P}\|y - z(t)\Lambda^{1/2}U^T\|_2^2. \tag{23}$$

In the oversampled setting ($P > N$), we define $\tilde{U} = (U, U^\perp) \in \mathbb{R}^{P \times P}$, and if $P \leq N$ we let $\tilde{U} = U$ so that in both cases $\tilde{U}\tilde{U}^T = I$. Rearranging the training error yields:

$$E_t(w(t)) = \frac{1}{P}\|y\tilde{U} - z(t)\Lambda^{1/2}U^T\tilde{U}\|_2^2 = \frac{1}{P}\|\tilde{\epsilon} + (\bar{z} - z(t))\Lambda^{1/2}\|_2^2 + \|\epsilon U^\perp\|_2^2. \tag{24}$$

Here $\tilde{\epsilon} = \epsilon U$. From (11), we derived that the learning dynamics in linear networks follow:

$$\tilde{\epsilon}_i + \sqrt{\lambda_i}(\bar{z}_i - z_i(t)) = \left(\sqrt{\lambda_i}(\bar{z}_i - z_i(0)) + \tilde{\epsilon}_i\right)e^{-\frac{\lambda_i t}{\tau}}. \tag{25}$$

Thus, we may write the training error as a function of time as:

$$E_t(w(t)) = \frac{1}{P}\left(\sum_{i=1}^{N}(\sqrt{\lambda_i}(\bar{z}_i - z_i(0)) + \tilde{\epsilon}_i)^2 e^{-\frac{2\lambda_i t}{\tau}} + \sum_{j=1}^{P-N}(\tilde{\epsilon}_j^\perp)^2\right), \tag{26}$$

where $\tilde{\epsilon}^\perp = \epsilon U^\perp$ and the second sum in the expression above equals zero if $P \leq N$. Note that the training error is strictly decaying with time. If $P < N$, the training error will approach zero as each data point is memorized. If we average the training error dynamics over the noise, parameter, and data distributions, we find:

$$\langle E_t(w(t))\rangle = \frac{1}{\alpha}\int \rho^{MP}(\lambda)\left(\lambda(\sigma_w^2 + (\sigma_w^0)^2) + \sigma_\epsilon^2\right)e^{-\frac{2\lambda t}{\tau}}d\lambda + \left(1 - \frac{1}{\alpha}\right)\sigma_\epsilon^2 \mathbf{1}[\alpha > 1]. \tag{27}$$

See Fig 3 for demonstration that the theoretical prediction above matches the results of simulations.

This formula for training error dynamics may help explain recent empirical findings (e.g. C. Zhang et al., 2017; Arpit et al., 2017) that neural networks do not find it difficult to memorize noisy labels and require only slightly longer training times to do so. If we consider the amount of training time before $E_t$ reaches some pre-determined small value, this will increase slightly as noise is added to the output labels. If we consider the undersampled setting in which noise can be memorized ($\alpha < 1$), then at very late times ($t \gg \frac{\tau}{\lambda_{\min}}$) the training error will decay exponentially, so that to a good approximation:

$$E_t(w(t)) \propto \left(\lambda_{\min} + \frac{1}{\text{SNR}}\right)e^{-\frac{2\lambda_{\min}t}{\tau}}. \tag{28}$$

It follows that the time required to reach a training error proportional to $\delta$ scales as:

$$t \approx \frac{\tau}{2\lambda_{\min}}\log\left(\frac{\lambda_{\min} + \frac{1}{\text{SNR}}}{\delta}\right). \tag{29}$$

Thus, increasing the variance in the output noise should lead to only a logarithmic increase in the time required to memorize a dataset.

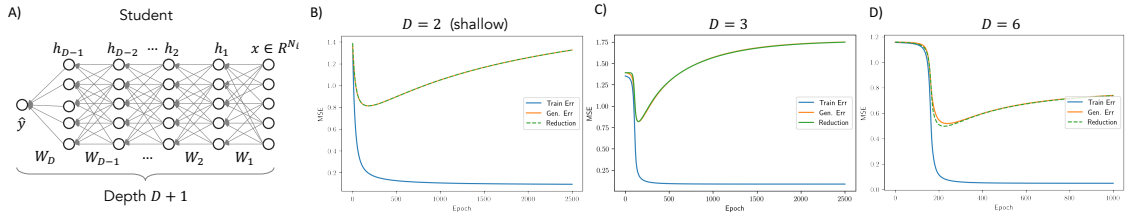

Figure 5: Reduction of deep linear network dynamics. (A) The student network is a deep linear network of depth $D+1$. (B-D) Comparisons of simulations of the full dynamics for networks initialized with small weights (yellow) to simulations of the reduced dynamics (green) for different depths and parameter settings.

## 3. Generalization dynamics in deep linear neural networks

We now turn to the impact of depth on generalization dynamics. We consider a deep linear student network (Fig. 5A) with weights $W_1, ..., W_D$ forming a $(D+1)$-layer linear network. In response to an input $x \in \mathbb{R}^N$, the student produces a scalar prediction $\hat{y} \in \mathbb{R}$ following the rule $\hat{y} = W_D W_{D-1} \cdots W_2 W_1 x = W^{\text{tot}} x$. We initially make no assumptions about the size of the layers, except for the final layer which produces a single scalar output. Continuous time gradient descent on the mean squared training error yields the dynamics,

$$\tau \frac{d}{dt} W_l = \left(\prod_{i=l+1}^{D} W_i\right)^T \left[\Sigma^{yx} - \left(\prod_{i=1}^{D} W_i\right) \Sigma^{xx}\right] \left(\prod_{i=1}^{l-1} W_i\right)^T. \tag{30}$$

These equations involve coupling between weights at different layers of the network, and up to cubic nonlinear interactions in the weights. While they compute a simple input-output map, deep linear networks retain several important features of nonlinear networks: most notably, the learning problem is nonconvex (P. Baldi and Hornik, 1989; Kawaguchi, 2016), yielding nonlinear dynamical trajectories (Fukumizu, 1998; Saxe, McClelland, and Ganguli, 2014).

We wish to find a simplified expression that provides a good description of these dynamics starting from small random weights. Intuitively, the full gradient dynamics contain repulsion forces that attempt to orthogonalize the hidden units, such that no two represent the same feature of the input. When initializing with small random weights, however, hidden units will typically already be nearly orthogonal–and hence these orthogonalizing repulsive forces can be safely neglected as in (Saxe, McClelland, and Ganguli, 2014). We begin by using an SVD-based change of variables to reduce coupling in these equations. We then will study the dynamics assuming that hidden units are initially *fully* decoupled–since this remains a good approximation for the full dynamics when starting from small random weights, as verified through simulation.

In particular, we make the following change of variables: $W_1(t) = r_2 z(t) V^T$, where $z(t) \in R^{1 \times N_i}$ is a row vector encoding the time-varying overlap with each principle axis in the input (recall $\Sigma^{xx} = V \Lambda V^T$); and $W_l(t) = d(t) r_{l+1} r_l^T, l = 2, \cdots, D$ where the vectors $r_i \in R^{N_l \times 1}$ are arbitrary unit norm vectors ($r_l^T r_l = 1$) specifying freedom in the internal representation of the network, and $d(t)$ is a scalar encoding the change in representation over time. With these definitions, $W^{\text{tot}} = \left(\prod_{l=2}^{D} d(t) r_{l+1} r_l^T\right) r_2 z(t) V^T = d(t)^{D-1} z(t) V^T = u(t) z(t) V^T$ where we have defined the scalar $u(t) = d(t)^{D-1}/(D-1)$. We show in Appendix A that solutions which start in this form remain in this form, and yield the following exact reduction of the dynamics:

$$\tau \dot{u} = u^{\frac{2D-4}{D-1}} \left(\tilde{s} z^T - u z \Lambda z^T\right), \tag{31}$$

$$\tau \dot{z} = \frac{1}{D-1} u \left(\tilde{s} - u z \Lambda\right). \tag{32}$$

These equations constitute a considerable simplification of the dynamics. Note that if there are $N_h = N_i$ hidden units in each layer, then full gradient descent in a depth $D+1$ network involves $O(N_i^2 D)$ parameters. The above reduction requires only $N_i + 1$ parameters, regardless of depth. Fig. 5 compares the predicted generalization dynamics from these reduced dynamics to simulations of full gradient descent for networks starting with small random weights and of different depths, confirming that the reduction provides an accurate picture of the full dynamics when starting from small random weights.

In the reduction, all modes are coupled together through the global scalar $u(t)$. Comparing (32) with (3), this scalar premultiplies the shallow dynamics, yielding a characteristic slow down early in training when both $z$ and $u$ are small (see initial plateaus in Fig. 5C-D). This behavior is a hallmark of deep dynamics initialized close to the saddle point where all weights are zero, and has been found for training error dynamics as well (Saxe, McClelland, and Ganguli, 2014; Goodfellow, Vinyals, and Saxe, 2015). Remarkably, in the reduction the entire effect of depth is compressed into the scalar $u$ which sets a global speed of the dynamics shared by all modes. Otherwise, each mode's dynamics is analogous to the shallow case and driven primarily by the size of its associated eigenvalue $\lambda_i$, with smaller eigenvalues being learned later. This suggests that optimal stopping will again be effective in deep networks, with comparable results to the shallow case. And as in the shallow case, eigenvalues that are zero make no progress, yielding a frozen subspace in which no learning occurs. Hence again, large initial weight norms will harm generalization error in the limited data regime in deep linear networks.

Thus for deep linear networks producing scalar outputs, our reduction predicts behavior qualitatively like their shallow counterparts in terms of their early stopping error and sensitivity to weight initializaitons. We note that extending our results to the multiple-output case is nontrivial and may yield important differences between shallow and deep generalization dynamics. Prior work on training dynamics, for instance, has shown that deep networks can exhibit multiple stage-like transitions to lower error over the course of training, while shallow networks cannot (Saxe, McClelland, and Ganguli, 2014). Unfortunately these results require mutually diagonalizable input and input-output correlations ($\Sigma^{yx} = USV^T$ and $\Sigma^{xx} = V\Lambda V^T$ for the same $V$), but even tasks which are mutually diagonalizable in the infinite data limit will typically not be on a finite batch of data, making these solutions unsuitable for investigating generalization error. Finally, the reduction points to a potential tension between training speed and generalization performance: small initial weights cause the network to be close to a saddle point which slows initial training progress, but large initial weights can remain indefinitely in the high dimensional regime, harming generalization. This may underly observations in full nonlinear networks that certain initializations

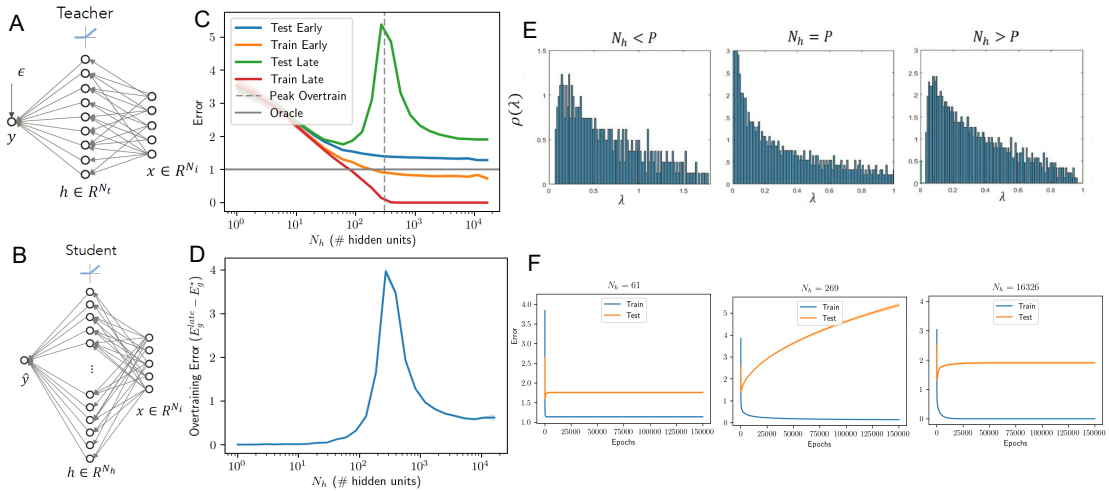

Figure 6: Learning from a nonlinear teacher. (A) The teacher network ($N_t$ ReLUs). (B) The student network ($N_h$ ReLUs, but only the output weights are trained). (C) Effect of model complexity. Optimal early stopping errors as a function of number of hidden units $N_h$ for the case $\text{SNR} = 1, \text{INR} = 0, N_i = 15, N_t = 30$ and $P = 300$ training samples. Shaded regions show $\pm 1$ standard error of the mean over 50 random seeds. (D) Overtraining peaks at an intermediate level of complexity near the number of training samples: when the number of free parameters in the student network equals the number of samples (300). (E) The eigenspectrum of the hidden layer of a random non-linear neural network with $P = 1000$ samples and an $N_i = 100$ dimensional input space. We consider three cases and find a similar eigenvalue density to a rescaled Marchenko-Pastur distribution when we concentrate only on the small eigenvalues and ignore a secondary cluster of $N_i$ eigenvalues farther from the origin. Left: Fewer hidden nodes than samples ($N_h = 500$ hidden units) leads to a gap near the origin and no zero eigenvalues. Center: An equal number of hidden nodes and samples ($N_h = 1000$) leads to no gap near the origin so that eigenvalues become more probable as we approach the origin. Right: More hidden nodes than samples ($N_h = 2000$) leads to a delta function spike of probability 0.5 at the origin with a gap to the next eigenvalue. (F) Average training dynamics for several models illustrating overtraining at intermediate levels of complexity.

can converge quickly but yield poorer generalization performance (e.g., Mishkin and Matas, 2016).

## 4. Generalization dynamics and optimal size in nonlinear neural networks

On practical problems, nonlinearity is essential since it allows networks to express complex functions. In this section we explore the degree to which the qualitative intuitions obtained through studying linear networks transfer to the nonlinear setting. Nonlinearity introduces the crucial question of model complexity: is there an optimal size of the student, given the complexity of the teacher and the number of training samples available? We again consider a student-teacher scenario, as depicted in Fig. 6A-B. The teacher is now a single hidden layer neural network with rectified linear unit (ReLU) activation functions and $N_t$ hidden neurons (see Seung, Sompolinsky, and Tishby, 1992; Watkin, Rau, and Biehl, 1993; Saad and Solla, 1995; Engel and Van Den Broeck, 2001; Wei, J. Zhang, et al., 2008; Wei and Amari, 2008 for similar approaches in other nonlinear networks). We draw the parameters of this network randomly. To account for the compression in variance of the ReLU nonlinearity, we scale the variance of the input-to-hidden weights by a factor $\sqrt{2}$ (Saxe, McClelland, and Ganguli, 2014; He et al., 2016). We generate a dataset as before by drawing $P$ random Gaussian inputs, obtaining the teacher's output $y$, and adding noise. The student network, itself a single hidden layer ReLU network with $N_h$ hidden units, is then trained to minimize the MSE on this dataset.

### 4.1 Trained output layer, fixed random first layer

We initially only train the second layer weights of the student network, leaving the first layer weights randomly drawn from the same distribution as the teacher. This setting most closely recreates the setting of the shallow linear case, as only the hidden-to-output weights of the student change during learning. Due to the nonlinearity, analytical results are challenging so we instead investigate the dynamics through simulations. We fixed the teacher parameters to SNR = 1, $N_t = 30$ and $N_i = 15$ and the number of training samples to $P = 300$, and then trained student models at a set of hidden layer sizes $N_h = [1, \cdots, 16326]$ ranging from $300\times$ smaller than the number of samples to $54\times$ larger. We average results at each size over 50 random seeds. Fig. 6C-D shows average errors as a function of the number of neurons in the student. Our goal is to understand whether the qualitative patterns of overtraining obtained from the linear analysis hold for this nonlinear case.

**Overtraining occurs at intermediate model complexity.** There is a peak in overtraining near where the number of samples in the dataset equals the number of hidden units/ free parameters, with larger models overtraining *less* than smaller models (Fig. 6C-D). Hence qualitatively, the network behaves akin to the linear case, with a peak in overtraining when the number of trainable parameters is matched to the number of training samples. To understand this behavior, we computed the eigenspectrum of the hidden layer activity in Fig. 6E. Despite the nonlinearity, we see qualitatively similar behavior to the linear case: the small eigenvalues approximately follow the Marchenko Pasteur distribution (c.f. Fig. 2). A similar observation about the small eigenvalues computed from the hidden layer was also used in a slightly different context to study training error in (Pennington and Bahri, 2017). We provide additional simulations in Fig. 9 showing this overtraining peak for different SNRs, which demonstrate that noisy data will amplify overtraining causing it to grow approximately proportionally to the inverse SNR as should be expected from (13).

**Larger models are better when early-stopped.** Strikingly, we find no evidence of overfitting: larger models are always better in terms of their early stopping test error. Here the teacher network has just 30 hidden neurons and the dataset consists of just 300 examples. Nevertheless, the best student model is the largest one tested, containing 16326 hidden units, or 344× as many hidden neurons as the teacher and 54× more neurons than training samples. This benefit of large models requires early stopping and initialization with small random weights, and hence reflects regularization through limiting the norm of the solution (see extended comments in Section 6). If instead training is continued to much longer times (green curve Fig. 6C), the optimal model size is smaller than the number of samples ($\approx 60$ hidden units), consistent with standard intuitions about model complexity.

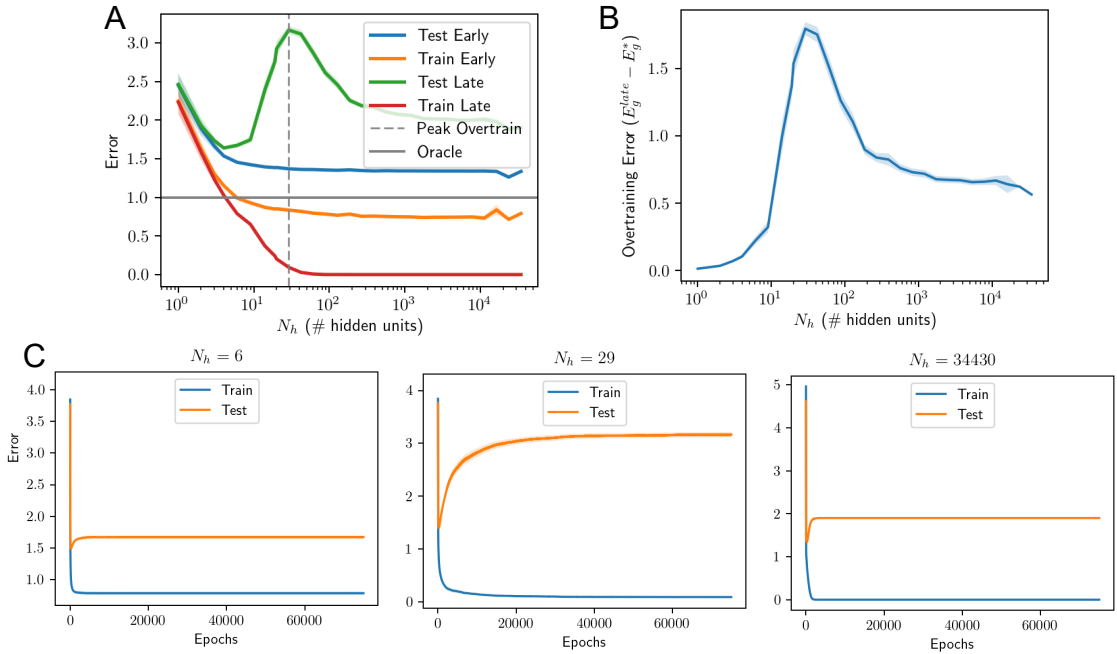

Figure 7: Training both layers of nonlinear student. These simulations are of the same setting as that of Fig 6, but with both layers of the student network trained. The number of hidden units for which the total number of free parameters is equal to the number of samples ($N_h + N_i N_h = P$) is marked with a dashed line and aligns well with the peak in overtraining observed.

### 4.2 Fully trained two-layer network

Next, we allowed both layers of the student network to train, as is typically the case in practice. Fig. 7 shows that similar dynamics are obtained in this setting. Overtraining is transient and peaked at intermediate levels of complexity, roughly where the number of free parameters ($N_i \cdot N_h + N_h$) equal the number of examples $P$. This is the point at which training error hits zero at long training times (red curve, Fig. 7A), consistent with recent results on the error landscape of ReLU networks (Soudry and Hoffer, 2017). Regarding model complexity, again we find that massive models relative to the size of the teacher and the number of samples are in fact optimal, if training is terminated correctly and the initial weights are small. In particular, good performance is obtained by a student with 34,430 hidden neurons, or 1147× the number of teacher neurons and 115× the number of examples. In contrast, without early stopping, the long-time test error has a unique minimum at a small model complexity of ≈ 5 hidden neurons in this instance. Hence as before, early stopping enables successful generalization in very large models, and the qualitative behavior is analogous to the linear case. We caution that the linear analysis can only be expected to hold approximately. As one example of an apparent difference in behavior, we note that the test error curve when the number of parameters equals the number of samples (Fig. 7C middle) diverges to infinity in the linear and fix hidden weights cases (see Fig. 2B middle and Fig. 6F middle), while it appears to asymptote in the two-layer trained case (see Fig. 7C middle). Nevertheless, this asymptote is much larger at the predicted peak (when the number of free parameters and samples are equal) than at other network sizes.

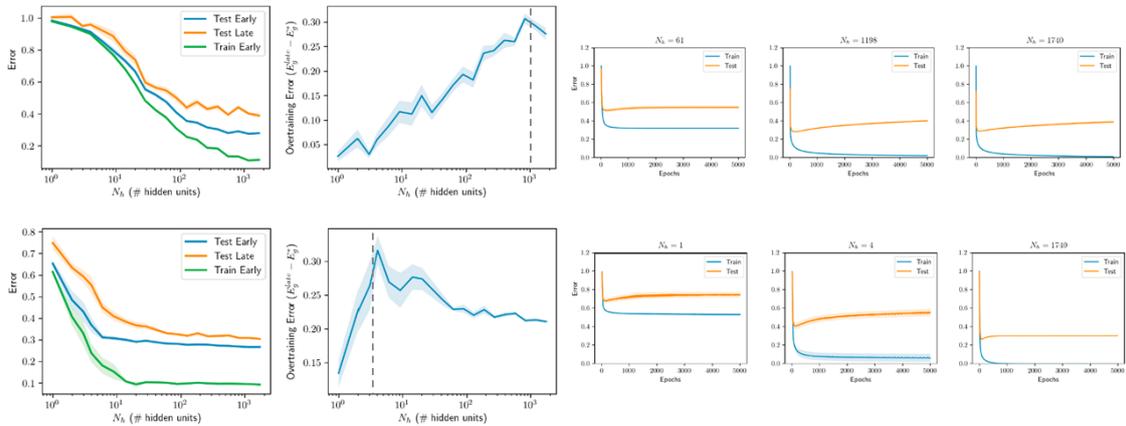

Figure 8: Dynamics of nonlinear networks on MNIST. Row 1: Network trained on binary MNIST classification of 7s vs 9s with 1024 training samples. First layer weights of student are random and fixed. Row 2: Same as Row 1 but with all student weights trained. In all cases the qualitative trends identified earlier hold: overtraining is a transient phenomenon at intermediate levels of complexity, and large models work well–no overfitting is observed given optimal early stopping. Note that the overtraining is less distinct than before primarily because these experiments were run for only 5000 epochs as opposed to tens of thousands of epochs as in Fig 6 and Fig 7, due to computational constraints.

### 4.3 Real-world datasets: MNIST

Finally, to investigate the degree to which our findings extend to real world datasets, we trained nonlinear networks on a binary digit recognition task using the MNIST dataset (LeCun, Cortes, and Burges, 1998). We trained networks to discriminate between $P = 1024$ images of 7s and 9s, following the setting of (Louart, Liao, and Couillet, 2017). Each input consists of a 28 × 28-pixel gray-scale image, flattened into a vector of length $N_i = 784$. Inputs were scaled to lie between 0 and 1 by dividing by 255 and each element was shifted to have zero mean across the training dataset. The target output was ± 1 depending on the class of the input, and the loss function remained the mean squared error on the training dataset. We train models using batch gradient descent with a single hidden layer for a fixed first layer and with both layers trained. Fig. 8 shows the resulting train and test errors for different model sizes and as a function of early stopping. Again, all qualitative features appear to hold: overtraining peaks when the number of parameters matches the number of samples, and larger models perform better overall. We note that in these experiments, due to computational restrictions, training was continued for only 5000 epochs of batch gradient descent (c.f. 75−150k epochs for the other experiments reported in this paper), yielding less pronounced overtraining peaks; and the maximum model size was 1740, making the peak in overtraining less identifiable, particularly for the fixed first layer case. There is additionally no evidence of overfitting, as the largest models outperformed smaller models. Overall, the qualitative findings accord well with those from the student-teacher paradigms, suggesting that our analysis accesses general behavior of generalization dynamics in gradient-trained neural networks.

## 5. Memorization and generalization

To generalize well, a model must identify the underlying rule behind a dataset rather than simply memorizing each training example in its particulars. An empirical approach to test for memorization is to see if a deep neural network can fit a training set with randomly scrambled labels rather than the true labels (C. Zhang et al., 2017). If a network can fit arbitrary random labels, it would seem to be able to memorize any arbitrary training dataset, and therefore, have excessive capacity and poor generalization performance. This approach has been taken by C. Zhang et al., 2017, and extended by Arpit et al., 2017. Their main empirical findings conflict with this intuitive story: while large deep networks do readily achieve zero training error on a dataset that has been randomly labeled, they nevertheless generalize well when trained on the true labels, even without any regularization or early stopping. Our results provide a straightforward explanation of these phenomena in terms of the signal-to-noise ratio of the dataset and the high dimensional dynamics of gradient descent.

In our student-teacher setting, a randomly-labeled training set corresponds to a situation where there is no rule linking input to output, and is realized in the regime where SNR → 0. As shown in Fig. 9A (red curve), nonlinear networks with more parameters than samples can easily attain zero training error on this pure-noise dataset. However, they do not generalize well after substantial training, or even after optimal early stopping, because the model has fit pure noise. When the exact same networks are instead trained on a high SNR dataset (Fig. 9B), they generalize extremely well, nearly saturating the oracle lower bound. Moreover, in the high-SNR regime, the performance gain from early stopping nearly disappears for larger network sizes. To take a specific example, the largest network size considered ($N_h = 16326$) can easily fit noise labels in the low-SNR regime, but attains the best generalization performance of all models in the high-SNR regime, even without any early stopping. This behavior arises due to two important phenomena that arise in the high-dimensional setting: First, in overcomplete networks, gradient descent dynamics do not move in the subspace where no data lies. This frozen subspace implicitly regularizes the complexity of the network when networks are initialized with small norm weights. Regardless of network size, only a $P-$dimensional subspace remains active in the learning dynamics. Second, as overcompleteness increases, the eigengap (smallest nonzero eigenvalue) of the hidden layer correlation matrix also increases due to the nature of the Marchenko Pasteur distribution, which protects against overtraining even at long training times. This increasing eigengap is a fundamental property of the high dimensional setting. Thus large neural networks can generalize well even without early stopping, provided that the rule to be learned in the dataset is prominent, the network is large, and the network is initialized with small weights. We provide more details on this finding through the lens of Rademacher complexity in the following section.

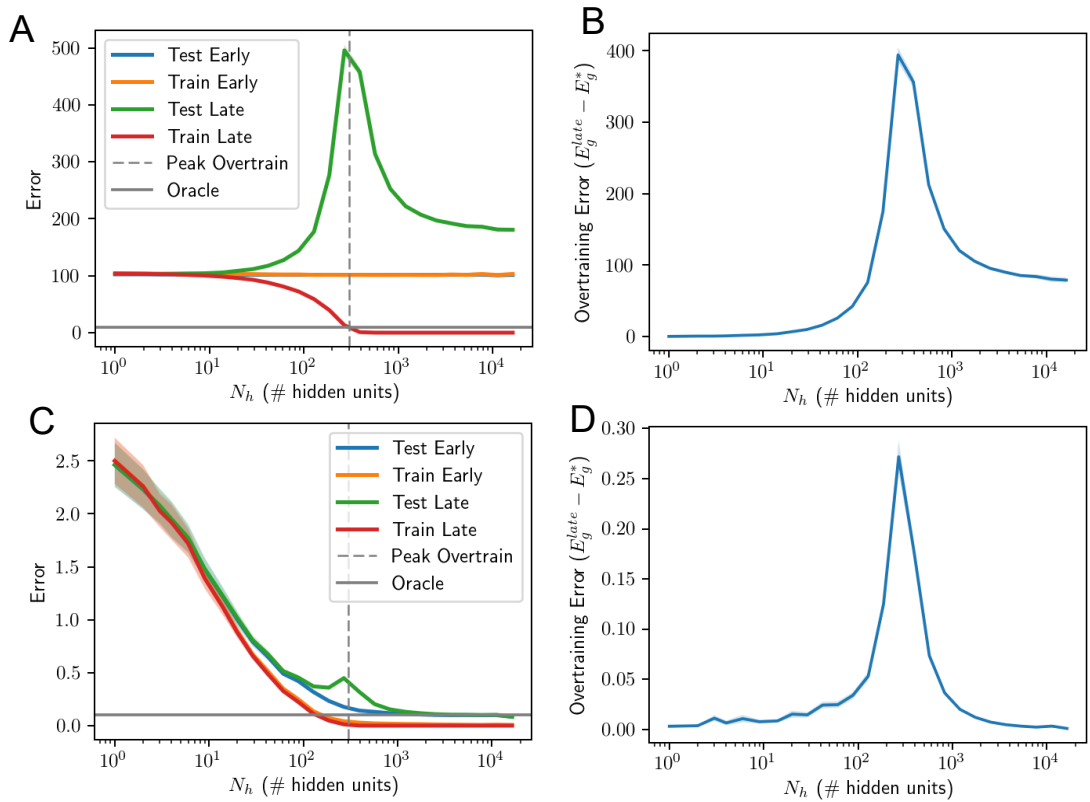

Figure 9: Memorization, generalization, and SNR. Nonlinear ReLU networks with fixed first layer (same setting as Fig 6) are trained on target labels of varying SNR. (A-B) Nearly random target labels (SNR = 0.01). In the high dimensional regime in which the model size outstrips the number of examples (here $N_h > 300$, indicated by vertical dotted line), all networks easily attain zero training error when trained long enough, thereby showing the ability to memorize an arbitrary dataset. (C-D) These exact same model sizes, however, generalize well when applied to data with nearly noise-free labels (SNR = 10). In fact in the low noise regime, even early stopping becomes unnecessary for large enough model sizes. The dynamics of gradient descent naturally protect generalization performance provided the network is initialized with small weights, because directions with no information have no dynamics.

## 6. Rademacher complexity and avoiding overfitting in non-linear neural networks

We want to understand why traditional Rademacher complexity bounds do not qualitatively match the overtraining picture observed in neural networks. In particular, they do not seem to square with the excellent generalization performance that can be achieved by large neural networks. To this end, we sketch how the dynamics of gradient descent learning may be included in the Rademacher complexity, yielding a bound which is more intuitively useful for understanding overtraining in neural networks. One particularly nice property of the bound we derive is that it shows how generalization can occur without early stopping: the Rademacher complexity can be expected to decrease with the number of hidden units $N_h$ for $N_h > P$ even when training is continued for long times.

As noted by C. Zhang et al., 2017, on the surface the excellent generalization ability of large networks seems to contradict the intuition from traditional measures such as Rademacher complexity and VC dimension. We find that a straightforward application of these measures yields not only loose bounds, but even the opposite qualitative behavior from what is observed in simulations.

A simple illustration of this can be seen in Fig. 6C where the task is fitting a random non-linear teacher. Here the size of the student network is increased well beyond the number of samples in the dataset and the number of hidden units in the teacher network, yet the generalization performance of early stopping continues to improve. The Rademacher complexity and VC dimension of the network is growing with the number of hidden units due to the fact that larger neural networks can more easily fit randomly labeled data, as discussed in C. Zhang et al., 2017. If we consider training a large non-linear neural network with a random first layer, the empirical Rademacher complexity of a classification problem (see e.g. Mohri, Rostamizadeh, and Talwalkar, 2012) is

$$\mathcal{R}(H) = \left\langle \sup_{h \in H} \frac{1}{P} \sum_{\mu=1}^{P} \sigma^\mu h(x^\mu) \right\rangle_{\sigma, x}, \tag{33}$$

where $H$ is the class of functions the network or classifier can have (mapping input $x^\mu$ to an output), and $\sigma^\mu = \pm 1$ are randomly chosen labels. The Rademacher complexity measures the ability of a network to learn random labels, and leads to a bound on the gap between the probability of correct classification of a new example or training example, denoted as generalization error and training error respectively below. The bound states that with probability at least $1 - \delta$,

$$E_g - E_t \leq 2\mathcal{R}(H) + \sqrt{\frac{\log \frac{1}{\delta}}{2P}}. \tag{34}$$

As we increase the width of a large random network, we see empirically that the number of points which may be fit exactly grows with the size of the network. Thus the bound above becomes trivial when $N_h > P$ since $\mathcal{R}(H) \geq O(1)$ in this limit and any random classification can be realized because solving for the weights in the final layer requires solving for $N_h$ unknowns with $P$ random constraint equations. This argument shows that the bound is loose, but this is to be expected from a worst-case analysis. More surprising is the difference in qualitative behavior: larger networks in fact do better than intermediate size networks at avoiding overtraining in our simulations.

However, we will see that a more careful bounding of the Rademacher complexity which uses the learning dynamics of wide networks does allow us to capture the overtraining behavior we observe in practice. For a rectified linear network with one hidden layer, a bound on the Rademacher complexity (Bartlett and Mendelson, 2002) is

$$\mathcal{R}(H) \leq \frac{B_2 B_1 C \sqrt{N_h}}{\sqrt{P}}, \tag{35}$$

where $B_2$ is the norm of the output weights, $B_1$ is the maximum L2 norm of the input weights to any hidden neuron, and $C$ is the maximum norm of the input data. From this it is clear that, no matter how many hidden neurons there are, the hypothesis complexity can be controlled by reducing the weight norm of the second layer.

Both early stopping and L2 regularization are capable of bounding the second layer weights and thus reducing overfitting, as we observe in simulations. In the case of early stopping, the dynamics scan through successively more expressive models with larger norms, and early stopping is used to pick an appropriate complexity.

Yet remarkably, even in the absence of early stopping we find that our analytic solutions provide a bound on the second layer weights in terms of the eigenspectrum of the hidden layer. The essential intuition is that, first, when a model is massively overcomplete, there is a large *frozen subspace* in which no learning occurs because no training samples lie in these directions; and second, the eigengap between the minimum eigenvalue and zero increases, protecting against overtraining and limiting the overall growth of the weights. To derive new bounds on the Rademacher complexity that account for these facts, we consider analytically how the weight norm depends on training time in our formulation. In the non-linear, fixed random first layer setting we are considering a model where

$$\hat{y}(x) = \sum_{a=1}^{N_h} \frac{w_a}{\sqrt{N_h}} \phi_a(x). \tag{36}$$

We will abuse notation because of the similarity with the single layer case, letting $X_h^{a\mu} = \frac{1}{\sqrt{N_h}} \phi_a(W_a^1 \cdot x^\mu)$. If we then diagonalize the covariance of the hidden layer s.t. $X_h X_h^T =$

$V_h \Lambda_h V_h^T$ and solve for the dynamics of the weights, we find:

$$z_i^h(t) = \frac{\tilde{z}_i^h}{\lambda_i^h}(1 - e^{-\frac{\lambda_i^h t}{\tau}}) + z_i^h(0)e^{-\frac{\lambda_i^h t}{\tau}}. \tag{37}$$

Here we define the vectors: $z^h = wV_h^T$, $\tilde{z}^h = yX_h^T V_h$, and $z_i^h(0) = w(0)V_h^T$, and we denote the eigenvalues of the hidden layer covariance as $\lambda_1^h, ..., \lambda_{N_h}^h$. Thus, the average squared norm of the output layer weights is

$$\|w(t)\|^2 = \sum_i z_i^2(t) = \sum_i \left( \frac{\|\tilde{z}_i^h\|^2}{(\lambda_i^h)^2}(1 - e^{-\frac{\lambda_i^h t}{\tau}})^2 + \|z_i^h(0)\|^2 e^{-2\frac{\lambda_i^h t}{\tau}} \right). \tag{38}$$

In the case of a linear shallow network, without a hidden layer, the expression for the growth of the norm of the weights is very similar, we simply substitute $\tilde{z}_i^h = \bar{z}_i^h + \frac{\tilde{\epsilon}_i}{\sqrt{\lambda_i^h}}$, and in this case the distribution of eigenvalues approach the Marchenko-Pasteur distribution in the high dimensional limit.

We now make the assumption that the initial weights are zero ($\sigma_w^0 = 0$), since this is the setting which will minimize the error and our large scale simulations are close to this limit (note that in deep networks we cannot set the initial weights exactly to zero because it would freeze the learning dynamics). It follows that $\langle \|w(t)\|^2 \rangle$ will monotonically increase as a function of time implying the Rademacher complexity bound is increasing with time.

The smallest non-zero eigenvalue constrains the maximum size that the norm of the weights can achieve, and zero eigenvalues result in no dynamics and thus do not impact the norm of the weights. Thus, even without early stopping, we can bound the norm of the weights in the hidden layer by an upper bound on the late-time behavior of (38):

$$\frac{\|w\|}{\sqrt{N_h}} \leq \sqrt{\frac{\max_i \|\tilde{z}_i^h\|^2}{\min_{i,\lambda_i^h > 0}(\lambda_i^h)^2} \frac{\min(P, N_h)}{N_h}} = B_2. \tag{39}$$

The minimum over $P$ and $N_h$ arises because there will be no impact on learning from eigenvalues of strength zero, corresponding to the frozen subspace. The presence of the minimum eigenvalue in the denominator indicates that, as the eigengap grows, the bound will improve. Substituting this into (35) yields a bound on the Rademacher complexity in terms of the eigenspectrum of the hidden layer:

$$\mathcal{R}(H) \leq B_1 C \sqrt{\frac{\max_i \|\tilde{z}_i^h\|^2}{\min_{i,\lambda_i^h > 0}(\lambda_i^h)^2} \frac{\min(P, N_h)}{P}}. \tag{40}$$

The above bound qualitatively matches our simulations with a non-linear network and fixed first layer weights, which show that the gap between training and generalization error at late stopping times drops as we increase the number of network parameters beyond the number of examples (see Fig. 6C-D) . In the equations above, as we increase $N_h$ above $P$, $\min(P, N_h)$ remains fixed at $P$, but we do increase the minimum non-zero eigenvalue of the hidden layer covariance matrix (see e.g. Fig 6E). This gap in the eigenspectrum can thus reduce the Rademacher complexity by bounding attainable network weights in massively overcomplete networks.

## 7. Discussion

Thus the dynamics of gradient descent learning in the high-dimensional regime conspire to yield acceptable generalization performance in spite of large model sizes. Making networks very large, even when they have more free parameters than the number of samples in a dataset, can reduce overtraining because many of the directions of the network have zero gradient and thus are never learned. This frozen subspace protects against overtraining regardless of whether learning is stopped early. The worst setting for overtraining is when the network width matches the number of samples in shallow networks and when the number of parameters matches the number of samples in nonlinear random networks. Thus, our analysis of the learning dynamics helps to explain why overtraining is not a severe problem in very large networks.

Additionally, we have shown that making a non-linear two layer network very large can continuously improve generalization performance both when the first layer is random and even when it is trained, despite the high Rademacher complexity of a deep network (C. Zhang et al., 2017). We demonstrate this effect both on learning from a random teacher and on an MNIST classification task.

Our findings result from random matrix theory predictions of a greater abundance of small eigenvalues when the number of samples matches the number of parameters. In the under- or over-complete regimes, learning is well-behaved due to the gap between the origin and the smallest non-zero eigenvalue of the input correlation matrix.

In our analysis, we have employed a simplified setting with Gaussian assumptions on the inputs, true parameters, and noise, and one might wonder whether our results can be expected to apply to more realistic scenarios. However, the Gaussian assumption on parameters and noise is not required for the analytic predictions on the generalization/training error dynamics of linear networks to hold. All that is required is that they are sampled *i.i.d.* with a finite mean and variance. In terms of other input distributions, we have shown empirically that our results apply for the MNIST dataset, but more broadly, there are theoretical reasons to believe these results to be somewhat robust: the Marchenko-Pasteur distribution is intuitively analogous to the law of large numbers. Just as a Gaussian distribution is the nearly universal limit of summing many independent (bounded moment) random variables, so the Marchenko-Pasteur distribution is a universal limit for the eigenvalues of random matrices containing *i.i.d.* elements drawn from non-Gaussian distributions with sub-exponential tails (see e.g. Pillai, Yin, et al., 2014). Thus, the predictions are theoretically justified when the noise, parameters, and input are selected *i.i.d.* from non-heavy tailed distributions. In the case of correlated input, our methods can be applied when the eigenspectrum of the input covariance can be computed or accurately estimated.

Finally, our analysis points to several factors under the control of practitioners that impact generalization performance. To make recommendations for practice, it is necessary to understand the regime in which high-performing deep networks typically operate. We suggest that the relevant regime is the high-dimensional, high-SNR setting ($\alpha < 1, \text{SNR} \gg 1$). In this regime, very large networks have dramatic advantages: they generalize better than their smaller counterparts, even without any regularization or early stopping (c.f. Fig. 9). Consider, for instance, our results on the MNIST dataset presented in Fig. 8. Here, while early stopping could improve performance, even at long training times the best model was the largest tested. For this setting, the practical message emerging from our theory is that larger models have no downside from a generalization perspective, provided they are initialized with small initial weights. Our results point to a strong impact of initialization on generalization error in the limited-data and large network regime: starting with small weights is essential to good generalization performance, regardless of whether a network is trained using early stopping.

Except for the deep linear network reduction, our results have focused on minimally deep networks with at most one hidden layer. It remains to be seen how these findings might generalize to deeper nonlinear networks (Kadmon and Sompolinsky, 2016), and if the requirement for good generalization (small weights) conflicts with the requirement for fast training speeds (large weights, Saxe, McClelland, and Ganguli, 2014) in very deep networks.

## Acknowledgments

We would like to thank Haim Sompolinsky, Kenyon Tsai, James Fitzgerald, and Pankaj Metha for many useful discussions as well as the Swartz Program in Theoretical Neuroscience at Harvard University for support.

## Appendix A. Deep linear neural network dynamics

Here we derive the reduction describing the dynamics of learning in deep linear neural networks. To begin, we make the following change of variables: $W_1(t) = r_2 z(t) V^T$, where $z(t) \in R^{1 \times N_i}$ is a vector encoding the time-varying overlap with each principle axis in the input (recall $\Sigma^{xx} = V\Lambda V^T$); and $W_l(t) = d(t) r_{l+1} r_l^T, l = 2, \cdots, D$ where the vectors $r_i \in R^{N_l \times 1}$ are arbitrary unit norm vectors ($r_l^T r_l = 1$) specifying freedom in the internal representation of the network, and $d(t)$ is a scalar encoding the change in representation over time. With these definitions, $W^{\text{tot}} = \left(\prod_{l=2}^{D} d(t) r_{l+1} r_l^T\right) r_2 z(t) V^T = d(t)^{D-1} z(t) V^T = u(t) z(t) V^T$ where we have defined the scalar $u(t) = d(t)^{D-1}$.

We thus have, for $l > 1$,

$$\tau \frac{d}{dt} W_l = \left(\prod_{i=l+1}^{D} W_i\right)^T \left[\Sigma^{yx} - \left(\prod_{i=1}^{D} W_i\right) \Sigma^{xx}\right] \left(\prod_{i=1}^{l-1} W_i\right)^T, \quad (41)$$

$$\tau \frac{d}{dt} d(t) r_{l+1} r_l^T = r_{l+1} d(t)^{D-l-1} \left[\Sigma^{yx} - d(t)^{D-1} z(t) V^T V \Lambda V^T\right] V d(t)^{l-1} z(t)^T r_l^T \quad (42)$$

$$\tau \frac{d}{dt} d(t) r_{l+1} r_l^T = d(t)^{D-2} \left[\tilde{s} z^T - d(t)^{D-1} z(t) \Lambda z(t)^T\right] r_{l+1} r_l^T \quad (43)$$

and equating coefficients yields the dynamics

$$\tau \frac{d}{dt} d(t) = d(t)^{D-2} \left[\tilde{s} z^T - d(t)^{D-1} z(t) \Lambda z(t)^T\right] \quad (44)$$

as desired. We then change variables to $u(t) = d(t)^{D-1}/(D-1)$ to derive the first differential equation:

$$\tau \dot{u} = u^{\frac{2D-4}{D-1}} \left(\tilde{s} z^T - u z \Lambda z^T\right). \quad (45)$$

Similarly for $l = 1$, we have

$$\tau \frac{d}{dt} W_1 = \left(\prod_{i=2}^{D} W_i\right)^T \left[\Sigma^{yx} - \left(\prod_{i=1}^{D} W_i\right) \Sigma^{xx}\right], \quad (46)$$

$$\tau \frac{d}{dt} r_2 z(t) V^T = r_2 d(t)^{D-1} \left[\Sigma^{yx} - d(t)^{D-1} z(t) V^T V \Lambda V^T\right], \quad (47)$$

$$\tau \frac{d}{dt} r_2 z(t) V^T = d(t)^{D-1} \left[\tilde{s} - d(t)^{D-1} z(t) \Lambda\right] V^T. \quad (48)$$

Again equating coefficients and making the substitution $u(t) = d(t)^{D-1}/(D-1)$ yields

$$\tau \dot{z} = \frac{1}{D-1} u \left(\tilde{s} - u z \Lambda\right). \quad (49)$$

## Appendix B. Optimal weight norm growth with SNR

Here we derive the optimal weight norm as a function of the signal to noise ratio ((20) in the main text). We analyze ridge regularized regression with strength $\gamma = \frac{1}{\text{SNR}}$,

$$\hat{w} = y X^T \left(X X^T + \frac{1}{\text{SNR}} I\right)^+. \quad (50)$$

Substituting from the model that
$$y = \bar{w}X + \epsilon, \tag{51}$$
yields:
$$\hat{w} = \left(\bar{w}XX^T + \epsilon X^T\right)\left(XX^T + \frac{1}{\text{SNR}}I\right)^+. \tag{52}$$
Utilizing the decompositions
$$XX^T = V\Lambda V^T, \tag{53}$$
$$X = V\Lambda^{1/2}U^T, \tag{54}$$
the previous equation can be written as
$$\hat{w}V = \left(\bar{w}V\Lambda + \epsilon U\sqrt{\Lambda}\right)(\Lambda + \gamma I)^{-1}. \tag{55}$$

Because the noise and parameter values are drawn iid, the rotations $U$ and $V$ will not impact their mean or moments, we can square the above equations and compute the average parameter strength:

$$\left\langle \hat{w}^2 \right\rangle = \sigma_w^2 \frac{1}{N}\sum_{i=1}^{N} \frac{\lambda_i}{\lambda_i + \frac{1}{\text{SNR}}} \to \sigma_w^2 \int \rho^{\text{MP}}(\lambda)\left(\frac{\lambda}{\lambda + \frac{1}{\text{SNR}}}\right). \tag{56}$$

Thus, we see that the average estimated parameter norm monotonically increases with SNR which makes sense intuitively because there is less regularization required as the data quality improves.